\documentclass[letterpaper]{article} % DO NOT CHANGE THIS
\usepackage{aaai2026}  % DO NOT CHANGE THIS
\usepackage{times}  % DO NOT CHANGE THIS
\usepackage{helvet}  % DO NOT CHANGE THIS
\usepackage{courier}  % DO NOT CHANGE THIS
\usepackage[hyphens]{url}  % DO NOT CHANGE THIS
\usepackage{graphicx} % DO NOT CHANGE THIS
\usepackage{natbib}  % DO NOT CHANGE THIS AND DO NOT ADD ANY OPTIONS TO IT
\usepackage{caption} % DO NOT CHANGE THIS AND DO NOT ADD ANY OPTIONS TO IT
\usepackage{algorithm}
\usepackage{algpseudocode}

\usepackage{newfloat}
\usepackage{listings}
\DeclareCaptionStyle{ruled}{labelfont=normalfont,labelsep=colon,strut=off} % DO NOT CHANGE THIS
\floatstyle{ruled}
\newfloat{listing}{tb}{lst}{}
\floatname{listing}{Listing}
\usepackage{amsmath,amssymb,amsfonts,amsthm}
\usepackage{epstopdf}
\usepackage{mathtools}
\usepackage{subcaption}
\usepackage{enumerate}
\usepackage[]{todonotes}

\DeclareFontFamily{U}{stix2bb}{}
\DeclareFontShape{U}{stix2bb}{m}{n} {<-> stix2-mathbb}{}
\NewDocumentCommand{\indicator}{}{\text{\usefont{U}{stix2bb}{m}{n}1}}

\newtheorem{theorem}{Theorem}
\newtheorem{lemma}{Lemma}
\newtheorem{corollary}{Corollary}
\newtheorem{definition}{Definition}  

\newtheorem{remark}{Remark}

\title{Inference Offloading for Cost-Sensitive Binary Classification at the Edge
}
\author{
    Vishnu Narayanan Moothedath\textsuperscript{\rm 1}, 
    Umang Agarwal\textsuperscript{\rm 2}, Umeshraja N\textsuperscript{\rm 2},\\
    James Richard Gross\textsuperscript{\rm 1}
    Jaya Prakash Champati\textsuperscript{\rm 3}
    Sharayu Moharir\textsuperscript{\rm 2}
}
\affiliations{
    \textsuperscript{\rm 1}Information Science and Engineering, KTH Royal Institute of Technology, Stockholm, Sweden\\
    \textsuperscript{\rm 2}Electrical Engineering, Indian Institute of Technology Bombay, Mumbai, India\\
    \textsuperscript{\rm 3}Computer Science, University of Victoria, Victoria, Canada.
}

\begin{document}

\maketitle

\begin{abstract}
We focus on a binary classification problem in an edge intelligence system where false negatives are more costly than false positives. The system has a compact, locally deployed model, which is supplemented by a larger, remote model, which is accessible via the network by incurring an offloading cost. For each sample, our system first uses the locally deployed model for inference. Based on the output of the local model, the sample may be offloaded to the remote model. This work aims to understand the fundamental trade-off between classification accuracy and the offloading costs within such a hierarchical inference (HI) system. To optimise this system, we propose an online learning framework that continuously adapts a pair of thresholds on the local model's confidence scores. These thresholds determine the prediction of the local model and whether a sample is classified locally or offloaded to the remote model. We present a closed-form solution for the setting where the local model is calibrated. For the more general case of uncalibrated models, we introduce H2T2, an online two-threshold hierarchical inference policy, and prove it achieves sublinear regret. H2T2 is model-agnostic, requires no training, and learns during the inference phase using limited feedback. Simulations on real-world datasets show that H2T2 consistently outperforms naive and single-threshold HI policies, sometimes even surpassing offline optima. The policy also demonstrates robustness to distribution shifts and adapts effectively to mismatched classifiers.
\end{abstract}

% Uncomment the following to link to your code, datasets, an extended version or similar.
% You must keep this block between (not within) the abstract and the main body of the paper.
% \begin{links}
%     \link{Code and datasets}{https://github.com/vnmo/H2T2-AAAI-2026}
%     \link{Extended version}{https://arxiv.org/abs/2509.15674}
% \end{links}

\begin{links}
\textbf{Publication Info }---\\{\quad This paper was presented at AAAI 2026 in Singapore.}
    \link{Code and datasets}{https://github.com/vnmo/H2T2-AAAI-2026}
    % \link{Published version}{https://arxiv.org/abs/2509.15674}
\end{links}

%%%%%%%%%%%%%%%%%%%%%%%%%%%%%%%%%%%%%%%%%%%%%%%%%%%%
%%%%%%%%%%%%%%%%%%%%%%%%%%%%%%%%%%%%%%%%%%%%%%%%%%%%
%%%%%%%%%%%%%%%%%%%%%%%%%%%%%%%%%%%%%%%%%%%%%%%%%%%%
%%%%%%%%%%%%%%%%%%%%%%%%%%%%%%%%%%%%%%%%%%%%%%%%%%%%
%%%%%%%%%%%%%%%%%%%%%%%%%%%%%%%%%%%%%%%%%%%%%%%%%%%%
%%%%%%%%%%%%%%%                      %%%%%%%%%%%%%%%
%%%%%%%%%%%%%%%        SECTION       %%%%%%%%%%%%%%%
%%%%%%%%%%%%%%%                      %%%%%%%%%%%%%%%
%%%%%%%%%%%%%%%%%%%%%%%%%%%%%%%%%%%%%%%%%%%%%%%%%%%%
%%%%%%%%%%%%%%%%%%%%%%%%%%%%%%%%%%%%%%%%%%%%%%%%%%%%
%%%%%%%%%%%%%%%%%%%%%%%%%%%%%%%%%%%%%%%%%%%%%%%%%%%%
%%%%%%%%%%%%%%%%%%%%%%%%%%%%%%%%%%%%%%%%%%%%%%%%%%%%
%%%%%%%%%%%%%%%%%%%%%%%%%%%%%%%%%%%%%%%%%%%%%%%%%%%%

\section{Introduction}
Edge intelligence systems often operate in critical and resource-constrained environments, where inference tasks must carefully balance cost and accuracy. Examples include event detection, typically framed as binary classification with applications such as medical diagnostics, surveillance, and smart cameras detecting theft, where high reliability must be achieved with low cost.
For instance, consider a medical scenario where a lightweight local inference model performs radiology screening of patient scans but may struggle with ambiguous or complex cases. Offloading such samples to a more accurate remote model--or a human specialist-- can considerably improve reliability, but by incurring significant networking and access costs.
The consequences of false positives (FP) and false negatives (FN) are typically asymmetric in such settings. For example, missing an incoming threat or a cancer case may be far costlier than a false alarm. 
Moreover, retraining or updating the local model to adapt to distribution shifts or cost changes may be impractical due to network limitations or model complexity. As a result, training-independent, inference-time decisions become essential.
In some systems, pre-training may not be possible at all, requiring models to learn on the fly~\cite{satyaSCML}.

\begin{figure}[t]
\centering
\includegraphics[width=\linewidth]{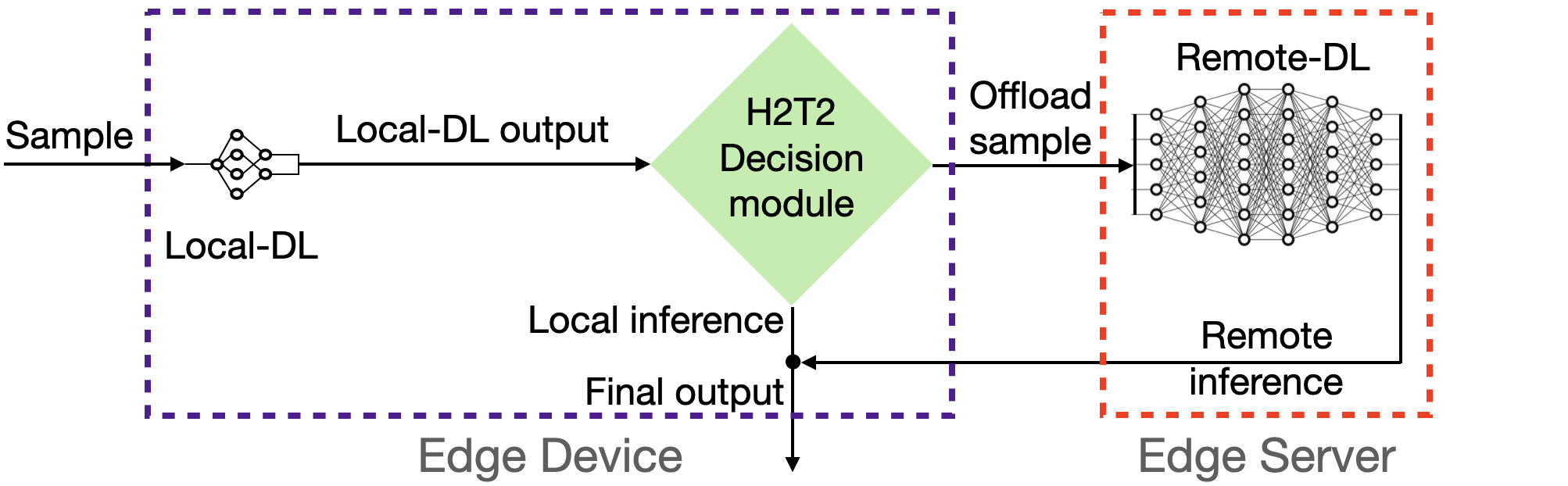}
\caption{Architecture of the proposed HI Hedge with Two Thresholds (H2T2) offload decision-making policy.} 
\label{fig:sysmodel}
\end{figure}

We consider an edge intelligence system with an edge device and a remote server. 
The device is equipped with a pre-trained local deep learning (LDL) model tasked with cost-sensitive binary classification. 
For each sample, it can either make a local decision using the LDL output or offload it--at a cost--to a larger, more accurate remote DL (RDL) model.
We aim to design an inference-time offloading policy with asymmetric FP and FN costs that selectively offload ambiguous samples based on the LDL output.
We use Hierarchical Inference (HI)~\cite{hiTMLCN,Beytur2024,hiMobihoc}, a post-hoc meta learning framework where each sample is first sent to the LDL, and the decision to offload to the RDL is made based on the LDL's output; see Figure \ref{fig:sysmodel}.
Prior HI works learned a threshold and offloads a sample if the LDL's confidence (e.g., maximum softmax value) of its decision falls below it.

\subsection{Related Works}
HI has received considerable attention~\cite{Nikoloska2021,hiSec,hiMobisys,hiTMLCN,Beytur2024, hiMobihoc,behera2025exploring,hiPMLR}, particularly within the edge computing community. 
\citeauthor{Nikoloska2021, hiMobisys,behera2025exploring} assumed a fixed offloading cost per sample and used a predetermined threshold--computed from the training data--on the maximum softmax value of the LDL to make the offloading decision. 
To handle the varying offloading costs and distribution drifts between training and inference phases, \citeauthor{hiTMLCN,Beytur2024,hiMobihoc,hiSec} investigated learning the optimum threshold online. 
\citeauthor{hiTMLCN} formulated HI as a variant of the classical Prediction with Expert Advice (PEA)~\cite{BianchiBook} and provided dataset-dependent regret bounds. 
\citeauthor{hiMobihoc} provided improved regret bounds for the problem, considering a more general setting with random and potentially correlated offloading costs. 
\citeauthor{Beytur2024} extended HI to multiple devices and studied maximising accuracy subject to an average offloading cost constraint. In contrast to existing work on HI, we consider asymmetric costs for FPs and FNs.

\begin{figure}[t]
    \centering   \includegraphics[width=\linewidth]{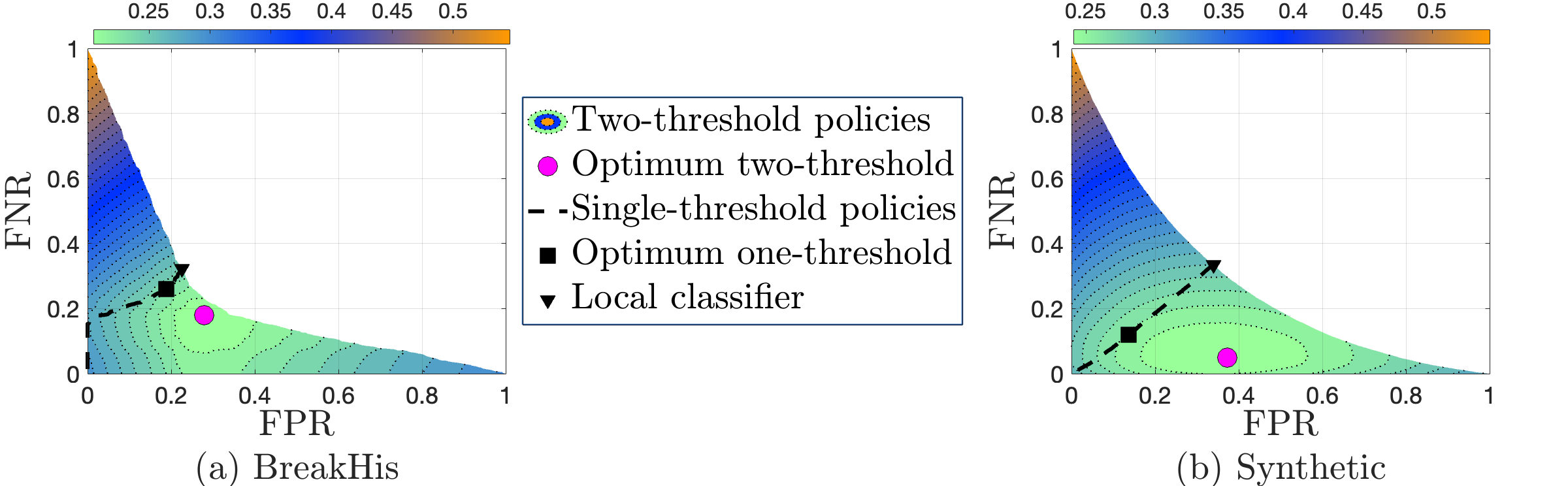}
    \caption{FPR vs. FNR and average cost of single- and two-threshold policies on  (a) BreakHis, and (b) synthetic configurations from TABLE~\ref{Table:Dataset}. Normalised costs of false positive, false negative, and offload are 0.7, 1, and 0.3, respectively.
   }
    \label{fig:fprvsfnr}
\end{figure}

Our work is also related to the Learning to Reject (L2R) and Learning to Defer (L2D) problems~\citeauthor{cortesLearningtoReject,saerens2002adjusting,ramaswamy2018consistent,ni2019calibration,narasimhan2022post}, which focus on training models with an option to reject, or defer to an expert model--at some cost.
In contrast, ours is a meta-learning framework proposed for the inference phase, where the pre-trained LDL is unaltered, and it does not require offline RDL inferences. This allows for the use of off-the-shelf models on the edge device and the server.  In contrast to cost-sensitive classification~\cite {elkan2001foundations}, our problem requires ambiguity-awareness not only for cost-sensitive classification but also to make the offloading decision. Finally, our framework handles Out-of-Distribution (OOD) samples at inference time, allows independent costs for local processing, offloading, false positives, and false negatives, and allows the offloading cost to be random and unknown a priori. We summarise the difference with related works in Table~\ref{Tab:RelatedworksComparison}.

While prior HI works use the LDL output to make naive \textit{argmax}-based LDL inference when not offloading, we use it to make our own cost-sensitive decisions--serving as both an offloading and a local inference policy. 
More importantly, all HI works use a single threshold for offloading decisions. We argue that, for cost-sensitive classification, using a single threshold results in sub-optimal solutions. To illustrate this, in Fig.~\ref{fig:fprvsfnr} we use two dataset-classifier combinations and show the false-positive rates (FPRs), false negative rates (FNRs), and the average costs achievable by single-threshold and two-threshold policies. While single-threshold policies can reduce FPR and FNR, even the best single-threshold policy cannot achieve the lowest possible overall cost. Instead, we propose a two-threshold HI decision rule and present a sublinear regret algorithm to solve the problem at hand.

\subsection{Our Contributions}
Our key contributions are outlined as follows:
\begin{itemize}
\item For calibrated LDL models, we derive a two-threshold rule for the softmax value that minimises the Bayesian expected cost. When FP and FN costs are equal and the offloading cost is constant, this rule reduces to Chow's rule for classification with rejection~\cite{chow2003optimum}. 
%\item Motivated by this, we formulate a two-threshold online learning problem for general uncalibrated LDL models.
\item Motivated by this, we propose a two-threshold-based HI policy called H2T2 to solve the problem and show that it has $O(T^\frac{2}{3})$ regret. Unlike the standard PEA, ground truth labels are unavailable during inference, and H2T2 achieves this sublinear regret by estimating local misclassifications via offloading a fraction of input samples.
\item Experiments on real-world and synthetic datasets demonstrate lower costs of H2T2 compared to the baseline policies, and in some cases, even offline single-threshold optima. While some cost regions favour naive policies, these regions are unknown a priori. We also show that H2T2 cost gains are significant for OOD data.
\end{itemize}
\begin{table}[t]
\centering
\setlength{\tabcolsep}{0.9mm}
\label{tab:relatedworks}
%\begin{tabular}{|l|c|c|c|c|c|c|}
\begin{tabular}{lcccccc}
\hline
Related work category           & {AA}          & {AC}          & {IT}      & DMA           \\ \hline   %\hline
Rejection(L2R), Deferral(L2D)   & \checkmark    & $\times$      &$\times$   & $\times$      \\ %\hline   
Cost-Sensitive Classification   & $\times$      & \checkmark    &$\times$   & $\times$      \\ %\hline   
HI--Single Threshold            & \checkmark    & $\times$      &\checkmark & \checkmark    \\ %\hline   
H2T2 (proposed)                 & \checkmark    & \checkmark    &\checkmark & \checkmark    \\ \hline   
\end{tabular}            
\caption{Comparison of the proposed policy with related work categories across key properties: ambiguity-aware (AA), asymmetric costs (AC), inference-time learning (ITL), and dataset \& model agnostic (DMA).}
\label{Tab:RelatedworksComparison}
\end{table}
In the following sections, we present the problem setup, theoretical results for calibrated models, the proposed algorithm with regret analysis, and experimental evaluation. Proofs are provided in the appendix.

%%%%%%%%%%%%%%%%%%%%%%%%%%%%%%%%%%%%%%%%%%%%%%%%%%%%
%%%%%%%%%%%%%%%%%%%%%%%%%%%%%%%%%%%%%%%%%%%%%%%%%%%%
%%%%%%%%%%%%%%%%%%%%%%%%%%%%%%%%%%%%%%%%%%%%%%%%%%%%
%%%%%%%%%%%%%%%%%%%%%%%%%%%%%%%%%%%%%%%%%%%%%%%%%%%%
%%%%%%%%%%%%%%%%%%%%%%%%%%%%%%%%%%%%%%%%%%%%%%%%%%%%
%%%%%%%%%%%%%%%                      %%%%%%%%%%%%%%%
%%%%%%%%%%%%%%%        SECTION       %%%%%%%%%%%%%%%
%%%%%%%%%%%%%%%                      %%%%%%%%%%%%%%%
%%%%%%%%%%%%%%%%%%%%%%%%%%%%%%%%%%%%%%%%%%%%%%%%%%%%
%%%%%%%%%%%%%%%%%%%%%%%%%%%%%%%%%%%%%%%%%%%%%%%%%%%%
%%%%%%%%%%%%%%%%%%%%%%%%%%%%%%%%%%%%%%%%%%%%%%%%%%%%
%%%%%%%%%%%%%%%%%%%%%%%%%%%%%%%%%%%%%%%%%%%%%%%%%%%%
%%%%%%%%%%%%%%%%%%%%%%%%%%%%%%%%%%%%%%%%%%%%%%%%%%%%

\section{Problem Setting}\label{problemSet}
We consider a system consisting of an edge device and an edge server. The device needs to infer relevant information from each sample arriving periodically, which is then used to make control decisions. 
This work focuses on the case where this inference task is a cost-sensitive binary classification or event detection.
The system is equipped with two DL (deep learning) models trained for this task: a local DL (LDL) deployed on the device, and a remote DL (RDL) deployed on the server. 
Given the memory and computing constraints of the device, the LDL is lightweight and, therefore, has lower accuracy than the RDL. 
We focus on the HI framework, where each sample is first processed by the LD. Based on its output, the system either makes a local inference or offloads the sample to the RDL for better inference.

\subsubsection{Inference Task}
Assume samples $x_t$ arrive in time slots indexed by $t$. 
The goal of the classifier is to map each sample to one of two possible classes, $0$ and $1$. 
To this end, the model outputs a softmax tuple $(f_{0}(x_t), f_{1}(x_t))\!\in\!\mathbb{R}^2$ with $0\!\leq\!f_{0}(x_t),\!f_{1}(x_t)\!\leq\!1$, and $f_{0}(x_t) \!+\!  f_{1}(x_t) \!= \!1$.
In the remainder of this work, we define the class $1$ as the class or event of interest and $f_t\coloneq f_{1}(x_t)$ as the corresponding LDL output.  We assume that the model outputs are expressed using $b$ bits and therefore are quantised into $2^b$ values. This is true for any practical system and has also been assumed in related literature.

\subsubsection{Loss Function}
We denote the LDL and RDL models by the binary classifiers $h_l(.)$ and $h_r(.)$. For each sample $x_t$, we either use $h_l(x_t)$ or offload the sample and use $h_r(x_t)$ as our inference.
In the absence of knowledge of the ground truth class for each sample, we treat the inference of the RDL as a proxy for this ground truth. This is justified by the fact that in any non-trivial scenario, the LDL aims to achieve the performance of the RDL, and no more. Further, this treatment adds simplicity and readability, and extensions to having an imperfect RDL can be done by following the same approach that is used for the single threshold HI in \citeauthor{hiTMLCN}.
Under this consideration, we have two types of classification errors that can occur when the inference is made locally:
\begin{enumerate}
    \item[--] False-positive (FP), when, $h_l(x_t) = 1$ and $h_r(x_t) = 0$,
    \item[--] False-negative (FN), when, $h_l(x_t) = 0$ and $h_r(x_t) = 1$.
\end{enumerate}
We use $\delta_1, \delta_{-1}$ and $\beta_t$ to denote the normalised false positive, false negative, and offload cost at time $t$. 
Here, $\beta_t \!<\! \beta$ is generated by an oblivious adversary, and is available at least after round $t$.
Normalisation is done by subtracting the common LDL processing cost and scaling by the maximum cost, mapping all values to $[0,1]$.
When a local inference is made, the system incurs a loss of $\phi_t$:
{\small\begin{align}
    \phi_t = 
\begin{cases}
\delta_1 & \text{if  FP}\\
\delta_{-1} & \text{if  FN}\\
0 & \text{otherwise.}
\end{cases} \label{eq:Phi_t}
\end{align}}
We assume that $\beta_t$ is presented to the policy at the start of round $t$. The system is trivial when $\beta_t\!\geq\!\max\{\delta_1,\delta_{-1}\}$, enabling an assumption of $\beta_t\!\leq\!\beta\!=\!1$.

Let $\pi$ denote the policy (or algorithm) making offloading decisions and $l_t(\pi)$ denote the incurred true loss of this policy at round $t$. It follows that
{\small\begin{align}
    l_t (\pi) = 
\begin{cases}
    \phi_t  & \text{if $h_l(x_t)$ is used} \\
    \beta_t &  \text{otherwise}.
\end{cases} \label{eq:lossFn}
\end{align}}
The corresponding cumulative loss $L_T(\pi)\!=\!\sum_{t=1}^{T} l_t(\pi)$. In the remainder of this work, we omit $\pi$ and refer to these as $l_t$ and $L_T$, for simplicity.

\subsection{Threshold-based Policies and Performance Metric}
Recall that for each sample, LDL outputs a score $f_t$ corresponding to class $1$. It follows that if $f_t$ is large (close to 1) or small (close to 0), it is very likely that the ground truth is 1 or 0.
On the other hand, intermediate values of $f_t$ imply ambiguity.
Motivated by this, we focus on threshold-based HI policies where the offloading decision in each slot $t$ is governed by a tuple $\Vec{\theta}\coloneq(\theta_l, \theta_{u})$, where $0\!\leq\!\theta_l\!\leq\!\theta_{u}\!\leq 1$. A sample is offloaded if $\theta_l\!\leq\!f_t \!<\!\theta_u$, predicted as class $1$ if $f_t\!\geq\!\theta_u$, and class $0$ otherwise.
Our goal in this work is to learn these thresholds online.

We say that a policy is a fixed-threshold (fixed-$\Vec{\theta}$) policy if $\Vec{\theta}$ does not change over the time-horizon of interest.
Let the losses associated with this threshold-based policy that uses $\Vec{\theta}$ as the threshold tuple be denoted by $\hat{l}_t(\Vec{\theta})$ and the corresponding cumulative loss be $\hat{L}_T(\Vec{\theta})$. That is, $\hat{l}_t(\Vec{\theta})={l}_t(\Vec{\theta})$ if the employed policy is a theshold-based policy. 
% For simplicity, we overload the notation $l_t$ from \eqref{eq:lossFn} and reuse it to denote $l_t(\Vec{\theta})$ and $L_t(\Vec{\theta})$, the instantaneous and cumulative losses incurred, respectively, by a fixed-threshold policy that uses $\Vec{\theta}$ as the threshold tuple. 
It follows that
{\small\begin{align}
    {\hat{l}_t(\Vec{\theta})} &= 
\begin{cases}
    % \beta & \text{if } \Vec{\theta},f_t\text{ results in offloading } x_t, \\
    \beta_t & \text{if } \theta_l\leq f_t<\theta_u, \\
    \phi_t & \text{otherwise.} 
\end{cases}\label{lossdefinition}
\end{align}}
Define $\Vec{\theta}^*\!=\!(\theta_l^*, \theta_{u}^*)$ as the fixed thresholds with the lowest cumulative loss in hindsight, in the class of fixed-$\vec{\theta}$ policies.
{\small\begin{align}
\Vec{\theta}^* &= \arg\min_{\Vec{\theta}} \; {\hat{L}_T(\Vec{\theta})} = \arg\min_{\Vec{\theta}} \; \sum_{t=1}^{T} {\hat{l}_t(\Vec{\theta})]}.\label{eq:thetastar}
\end{align}}
We use the fixed threshold policy with threshold $\Vec{\theta}^*$ as the reference policy to compare the performance of candidate policies. Our performance metric for a candidate policy is regret, denoted by $R_T$, where
{\small\begin{align}
R_T &= \mathbb{E}[L_T] - {\hat{L}_T(\Vec{\theta}^*)}.
    \label{reg_def}
\end{align}}
The expectation in \eqref{eq:thetastar} and \eqref{reg_def} are taken over the randomness in the candidate policy. Note that this notion of regret taking over the best fixed-threshold policy in the given input sequence is the stronger version where the regret, if bounded, holds over any input sequence, in a minimax sense. A weaker version of regret also exist, where $\Vec{\theta^*}$ is defined as the minimum in expectations.

%%%%%%%%%%%%%%%%%%%%%%%%%%%%%%%%%%%%%%%%%%%%%%%%%%%%
%%%%%%%%%%%%%%%%%%%%%%%%%%%%%%%%%%%%%%%%%%%%%%%%%%%%
%%%%%%%%%%%%%%%%%%%%%%%%%%%%%%%%%%%%%%%%%%%%%%%%%%%%
%%%%%%%%%%%%%%%%%%%%%%%%%%%%%%%%%%%%%%%%%%%%%%%%%%%%
%%%%%%%%%%%%%%%%%%%%%%%%%%%%%%%%%%%%%%%%%%%%%%%%%%%%
%%%%%%%%%%%%%%%                      %%%%%%%%%%%%%%%
%%%%%%%%%%%%%%%        SECTION       %%%%%%%%%%%%%%%
%%%%%%%%%%%%%%%                      %%%%%%%%%%%%%%%
%%%%%%%%%%%%%%%%%%%%%%%%%%%%%%%%%%%%%%%%%%%%%%%%%%%%
%%%%%%%%%%%%%%%%%%%%%%%%%%%%%%%%%%%%%%%%%%%%%%%%%%%%
%%%%%%%%%%%%%%%%%%%%%%%%%%%%%%%%%%%%%%%%%%%%%%%%%%%%
%%%%%%%%%%%%%%%%%%%%%%%%%%%%%%%%%%%%%%%%%%%%%%%%%%%%
%%%%%%%%%%%%%%%%%%%%%%%%%%%%%%%%%%%%%%%%%%%%%%%%%%%%

\section{Optimal Policy Under a Calibrated Model}\label{sec:calibrated}
In this section, we focus on the setting where the LDL is calibrated, i.e., for each sample, the LDL output (softmax value) for each class is equal to the a posteriori probabilities of that sample belonging to that class. 
Let $\cal{X}$ and $\cal{Y}$ denote the input space and the label space, respectively.
Assume that $(x_t,h_r(x_t))$ is drawn from $\cal{X}\times \cal{Y}$.
\begin{definition} [Calibrated Model]
\label{defn:cal} 
The DL model that generates the softmax values $(1-f_t, f_t)$ for the sample $x_t$ is calibrated, if
$\mathbb{P}(h_r(x_t) = 1 | x_t) = f_t,\,\forall t.$
\end{definition}
We now characterise the optimal HI offloading policy for cost-sensitive classification for calibrated LDL. 
\begin{theorem}
\label{thm:optimal_calibrated}
Let $x_t$ denote the sample input to a calibrated LDL. Then, the optimal predictor $h_l^*(x_t)$ is given by
{\small\begin{align}
 h_l^*(x_t) =
\begin{cases} 
1, & \text{if } f_t \geq \tfrac{\delta_1}{\delta_1 + \delta_{-1}}, \\
0, & \text{otherwise}.
\end{cases}\label{eq:Thm1_OptPredictor}
\end{align}}
Further,  with a known $\beta_t$, the optimum offloading decision for this setup is to offload the sample when
{\small\begin{align}
{\theta_l^*}^{(t)}\coloneq\tfrac{\beta_t}{\delta_{-1}}\leq f_t<1-\tfrac{\beta_t}{\delta_{1}}\coloneq{\theta_u^*}^{(t)},\label{eq:Thm1_thresholds}
\end{align}}
with an associated expected loss of 
{\small\begin{align}
    \mathbb{E}[\hat{l}_t]=\min\{\beta_t,\delta_{1}(1-f_t),\delta_{-1}f_t\}.\label{eq:Thm1_expectedLoss}
\end{align}}
\end{theorem}
\begin{remark}
Theorem~\ref{thm:optimal_calibrated} implies the following: 
(i) Offloading does not occur if $\beta_t \!\geq\! \tfrac{\delta_1 \delta_{-1}}{\delta_1 + \delta_{-1}}$, half the harmonic mean of $\delta_1$ and $\delta_{-1}$; 
(ii) when $\delta_1\!\!=\!\!\delta_{-1}$, the decision is equivalent to Chow's rule for classification with rejection, with no offloading if $\beta_t \!\geq\! 0.5$, and offloading if and only if $\min\{f_1(x_t), f_0(x_t)\}\!>\!\beta_t$. If not offloaded, the prediction is the class with the higher softmax value.
\end{remark}

The key question is its extension to non-calibrated models, where the thresholds must be learned.

%%%%%%%%%%%%%%%%%%%%%%%%%%%%%%%%%%%%%%%%%%%%%%%%%%%%
%%%%%%%%%%%%%%%%%%%%%%%%%%%%%%%%%%%%%%%%%%%%%%%%%%%%
%%%%%%%%%%%%%%%%%%%%%%%%%%%%%%%%%%%%%%%%%%%%%%%%%%%%
%%%%%%%%%%%%%%%%%%%%%%%%%%%%%%%%%%%%%%%%%%%%%%%%%%%%
%%%%%%%%%%%%%%%%%%%%%%%%%%%%%%%%%%%%%%%%%%%%%%%%%%%%
%%%%%%%%%%%%%%%                      %%%%%%%%%%%%%%%
%%%%%%%%%%%%%%%        SECTION       %%%%%%%%%%%%%%%
%%%%%%%%%%%%%%%                      %%%%%%%%%%%%%%%
%%%%%%%%%%%%%%%%%%%%%%%%%%%%%%%%%%%%%%%%%%%%%%%%%%%%
%%%%%%%%%%%%%%%%%%%%%%%%%%%%%%%%%%%%%%%%%%%%%%%%%%%%
%%%%%%%%%%%%%%%%%%%%%%%%%%%%%%%%%%%%%%%%%%%%%%%%%%%%
%%%%%%%%%%%%%%%%%%%%%%%%%%%%%%%%%%%%%%%%%%%%%%%%%%%%
%%%%%%%%%%%%%%%%%%%%%%%%%%%%%%%%%%%%%%%%%%%%%%%%%%%%

\section{The H2T2 Policy}\label{Sec:noncalibrated}
This section relaxes the assumption of a calibrated LDL, proposing instead an online policy that learns the values of the two thresholds. Any policy within our framework must make two decisions: first, whether to offload a given sample; and second, if the sample is not offloaded, the policy has to predict a class based on the LDL's output.

\subsubsection{Algorithm} 
\begin{figure}[t]
\centering
\includegraphics[width=\linewidth]{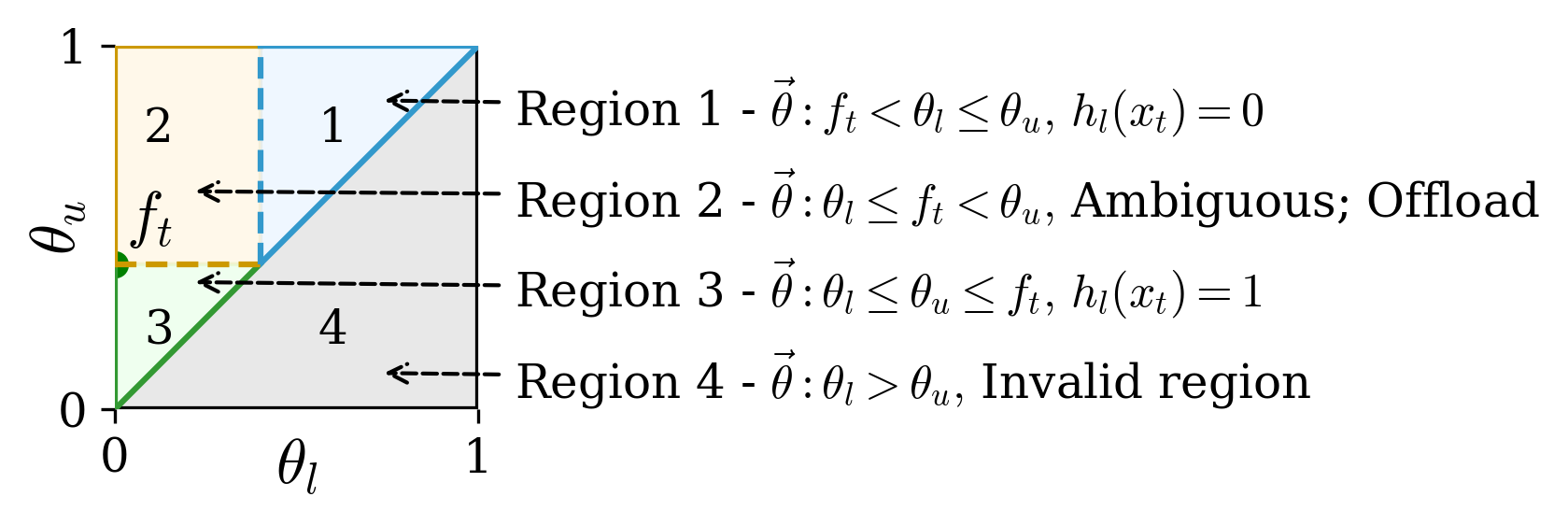}
\caption{Illustration showing the invalid and three valid regions of experts from \eqref{eqn:LDLpredictor} with respect to an observed $f_t$.
} 
\label{fig:illustration}
\end{figure}

We present the proposed HI-Hedge with Two Thresholds (H2T2) policy in Algorithm~\ref{alg:hedge_hi}. We note that H2T2 maps to the Prediction with Expert Advice (PEA) problem, where the \textit{experts} are the threshold tuples $\vec{\theta}$, and the expert chosen in round $t$ and $\vec{\theta}^{(t)}\coloneq(\theta_l^{(t)},\theta_u^{(t)})$. We denote the set of experts as $\Theta$. Given $\theta_l \!\leq\! \theta_u$, it follows that $|\Theta| \!=\! 2^{b-1}(2^{b}+1)$ experts.
Fig.~\ref{fig:illustration} illustrates three regions of the experts with respect to an observed $f_t$. 
We maintain weights for all experts. The probability of choosing an expert in a round is proportional to its weight in that round. This process will be discussed shortly. 
\begin{algorithm}[t]
\caption{H2T2: The two threshold HI policy.}
\label{alg:hedge_hi}
\begin{algorithmic}[1]
    \State $\epsilon$, $\eta$, $\Theta$.\Comment{\footnotesize Input\normalsize}
    \State $w_1(\Vec{\theta}) = 1 \hspace{3pt} \forall \hspace{3pt} \Vec{\theta} \in \Theta$, and $W_1 = |\Theta|$.\Comment{\footnotesize Initialization\normalsize}
    \For{$t \geq1$}
        \State Observe $f_t.$\Comment{\footnotesize LDL inference\normalsize}
        \State $p_t = \sum_{\theta_l< f_t}\sum_{\theta_u<f_t}w_t(\Vec{\theta}).$\Comment{\footnotesize Region 3\normalsize}
        \State $q_t = \sum_{\theta_l\leq f_t}\sum_{\theta_u>f_t}w_t(\Vec{\theta}).$\Comment{\footnotesize Region 2\normalsize}
        \State $\psi_t \sim \text{Uniform}(0,1)$.
        \State $\zeta_t \sim \text{Bernoulli}(\epsilon)$.%\Comment{Exploration RV} 
        \If{$\psi_t\leq q_t$ or $\zeta_t=1$}\Comment{\footnotesize Region 2, or exploration\normalsize}
            \State Offload $x_t$, receive $h_r(x_t)$, and observe $\phi_t$.
        \Else\Comment{\footnotesize Local inference\normalsize}
            \If{$\psi_t\leq q_t+p_t$} 
                \State $h_l(x_t)=1.$\Comment{\footnotesize Region 3\normalsize}
            \Else
                \State $h_l(x_t)=0.$\Comment{\footnotesize Region 1\normalsize}
            \EndIf
        \EndIf
        \ForAll{$\Vec{\theta} \in \Theta$}
                \State Compute $\tilde{l}_t(\Vec{\theta})$ using \eqref{eq:l_hat}.
                \State $w_{t+1}(\Vec{\theta}) = e^{-\eta \tilde{l}_t(\Vec{\theta})} w_t(\Vec{\theta})$.\Comment{\footnotesize Weight updation\normalsize}
            \EndFor
            \State $W_{t+1} = \sum_{\Vec{\theta} \in \Theta} w_{t+1}(\Vec{\theta})$.
    \EndFor
\end{algorithmic}
\end{algorithm}
\begin{table*}[t]
\centering
% \setlength{\tabcolsep}{0.7mm}
%\begin{tabular}{|l|l|l|l|l|l|}
\begin{tabular}{llllll}
\hline
Dataset            &Test Size  &LDL            &Accuracy       & FP      & FN    \\   \hline  
BreakHis       &$3365$   &MobileNet-based       &72\%           & 10\%    & 18\%   \\ %\hline
Chest          &$278$    &MobileNet-based       &64\%           & 16\%    & 20\%   \\ %\hline
Phishing       &$1106$   &Logistic regression   &75\%           & 12\%    & 13\%   \\ %\hline
ChestXRay      &$624$    &MobileNet-based       &78\%           & 18\%    & 4\%   \\ %\hline
% Synthetic      &$10^5$   & Naive                &66\%           & 15\%    & 19\%   \\ %\hline
ResnetDogs    &$2000$   &ResNet                &73\%           & 15\%    & 11\%   \\ %\hline
LogisticDogs  &$2000$   & Logistic regression  &56\%           & 22\%    & 22\%   \\ %\hline
BreaCh$^\dagger$&$7909$   &CNN                   &45\%$^\dagger$  & 17\%    & 38\%   \\
X-RaCT$^\dagger$&$5856$   &Chest classifier      &35\%$^\dagger$  & 1\%    & 64\%   \\ \hline
\end{tabular}
\caption{Different dataset-model pairs used in this technical appendix. $^\dagger$Note that BreaCh and X-RaCT are OOD datasets, and the accuracy below 50\% is intentional.}
\label{Table:Dataset}
\end{table*}
Under H2T2, given the output of the LDL and the expert (i.e., threshold tuple) chosen in round $t$, denoted with superscript $^{(t)}$, the local prediction $h_l(x_t)$ is defined as follows:
{\small\begin{align}
 h_l(x_t) =
\begin{cases} 
0, & \text{if } f_t<\theta_l^{(t)}\leq\theta_u^{(t)}, \\
\text{Ambiguous}, &\text{if } \theta_l^{(t)}\leq f_t<\theta_u^{(t)},\\
1, & \text{if } \theta_l^{(t)}\leq\theta_u^{(t)}\leq f_t.
\end{cases}\label{eqn:LDLpredictor}
\end{align}}

H2T2 offloads all samples for which $h_l(x_t)$ is ambiguous.
Otherwise, it still offloads with a probability $\epsilon$ based on a Bernoulli random variable $\zeta_t\!\sim\!\text{Ber}(\epsilon)$. The motivation of this \textit{exploration} step will be explained shortly. 
If H2T2 does not offload, $h_l(x_t)$ is the predicted class. 

We now discuss the process of computing the weights under H2T2.
Recall that, in our setting, we use the inference of the RDL as a proxy for the ground truth. One of the key challenges in our setting is that this inference is observed only when we offload a sample, and therefore, unlike the classical PEA setting, we learn the loss incurred by the experts only when we offload a sample.

To address this challenge, H2T2 employs two steps: 
First is the exploration step discussed earlier, where unambiguous samples are randomly offloaded with a probability $\epsilon$.
Next, H2T2 maintains an estimated loss $\tilde{l}_t(\Vec{\theta})$ for each expert that is used for weight updation.
Let $\indicator_{x_t}(\Vec{\theta})$ be an indicator random variable that is one if $h_l(x_t)$ would have been ambiguous if the expert chosen in time $t$ was $\Vec{\theta}$,\; i.e., $\theta_l\!\leq\!f_t\!<\!\theta_u$. 

 First we write the true loss from \eqref{eq:lossFn}, under H2T2:
{\small\begin{align}
    {l}_t(\Vec{\theta}^{(t)})\!=\!
\begin{cases} 
{\phi_t}    &\text{if }          \indicator_{x_t}(\Vec{\theta}^{(t)})\!=\!0, \zeta_t=0,
% \quad\left(\text{\textit{Offload}}\right
\\
\beta_t &\text{otherwise.} 
% \qquad\qquad\quad\;\left(\text{$h_t(x_t)$ is used}\right)
                            \end{cases}
\label{eq:h2t2trueloss}
\end{align}}
For simplicity, We drop the superscript $\cdot^{(t)}$ here onward, when the time index is already implied. 
Now, we define the estimated loss:
% {\small\begin{align}
%     \tilde{l}_t(\Vec{\theta})\!=\!
% \begin{cases} 
% \beta_t                     &\text{if }          O_t\!=\!1,\indicator_{x_t}(\Vec{\theta})\!=\!1,\\
% \tfrac{\phi_t}{\epsilon}    &\text{if }          E_t\!=\!1,\indicator_{x_t}(\Vec{\theta})\!=\!0 \\
% 0                           &\text{otherwise.}           \\
%                             \end{cases}
% \label{eq:l_hat}
% \end{align}}
{\small\begin{align}
    \tilde{l}_t(\Vec{\theta})\!=\!
\begin{cases} 
\beta_t                     &\text{if }          \indicator_{x_t}(\Vec{\theta})\!=\!1,\\
\tfrac{\phi_t}{\epsilon}    &\text{if }          \indicator_{x_t}(\Vec{\theta})\!=\!0, \zeta_t=1\\
0                           &\text{otherwise.}          
                            \end{cases}
\label{eq:l_hat}
\end{align}}
Also, let $\tilde{L}_T\!=\!\sum_{t=1}^T\tilde{l}_t$ be the corresponding cumulative estimated loss. 
Each expert’s weight in a round is an exponential function of the cumulative estimated loss of that expert up to that round, with a learning rate $\eta$. 
 Note that this estimated loss $\tilde{l}_t$ is used as part of the H2T2 policy for weight updation and is different from the true loss ${l}_t$ incurred by the algorithm. 
%The true loss incurred by the algorithm depends only on the offloading decisions and is given in \eqref{eq:lossFn}.

In our setting, all experts that yield the same $h_l(x_t)$ create contiguous and convex regions, as illustrated in Fig.~\ref{fig:illustration}. This simplifies the decision-making process: it suffices to compute the total probability of all experts in each of the three regions to decide whether to offload and to determine the inference of the LDL.

\subsubsection{Regret analysis} We now provide regret guarantees for H2T2. We first state the following two lemmas. 
Lemma~\ref{lemma:unbiased} states that $\tilde{l}_t(\Vec{\theta})$ is an unbiased estimator of the  losses incurred by a threshold-based policy {$\hat{l}_t(\Vec{\theta})$ in \eqref{lossdefinition}}, used for regret computation.
\begin{lemma}
\label{lemma:unbiased}
For all values of $f_t$ and for all $\Vec{\theta} \in \Theta$:
{\small\begin{align*}
    \mathbb{E}_{\zeta}[\tilde{l}_t(\Vec{\theta})] = {\hat{l}_t(\Vec{\theta})}. 
\end{align*}}
\end{lemma}
Next, in Lemma~\ref{lemma:expDifference}, we bound the difference in the expected values of the cumulative true loss (see \eqref{eq:h2t2trueloss}) and the cumulative loss of the fixed threshold policy, where the expectation is over the randomness of the policy, that is over both the randomness of the expert selection and the exploration with $\zeta$.
\begin{lemma}\label{lemma:expDifference}
Under H2T2 and a time horizon $T$:
{\small\begin{align*}
    \mathbb{E}\left[ L_T \right] - \mathbb{E} \left[{\hat{L}_T}\right]\leq\epsilon\beta T \leq \epsilon T.
\end{align*}}
\end{lemma}
\noindent We use these lemmas to upper bound the regret of H2T2.
\begin{theorem}\label{thm:regret}
The regret $R_T$ of H2T2 satisfies
{\small\begin{align}
    R_T \leq \left(\epsilon\beta  + \tfrac{\eta}{2\epsilon} \right) T + \tfrac{\ln(|\Theta|)}{\eta}. 
\end{align}}
\end{theorem}
\noindent The regret-minimizing values of $\eta$ and $\epsilon$ can be obtained by setting the partial derivatives of $R_T$ to zero.
\begin{corollary}\label{corol2}
$R_T$ is minimized by $\epsilon^*\!=\!\left({\ln(|\Theta|)}/{2\beta^2T}\right)^{{1}/{3}}$ and $\eta^*=\sqrt{{2\epsilon^*\ln{|\Theta|}}/{T}}.$ H2T2 with $\eta^*$ and $\epsilon^*$ has $O\left(T^{{2}/{3}}\right)$ regret.
\end{corollary}
\noindent We thus note that H2T2 has sublinear regret.

%%%%%%%%%%%%%%%%%%%%%%%%%%%%%%%%%%%%%%%%%%%%%%%%%%%%
%%%%%%%%%%%%%%%%%%%%%%%%%%%%%%%%%%%%%%%%%%%%%%%%%%%%
%%%%%%%%%%%%%%%%%%%%%%%%%%%%%%%%%%%%%%%%%%%%%%%%%%%%
%%%%%%%%%%%%%%%%%%%%%%%%%%%%%%%%%%%%%%%%%%%%%%%%%%%%
%%%%%%%%%%%%%%%%%%%%%%%%%%%%%%%%%%%%%%%%%%%%%%%%%%%%
%%%%%%%%%%%%%%%                      %%%%%%%%%%%%%%%
%%%%%%%%%%%%%%%        SECTION       %%%%%%%%%%%%%%%
%%%%%%%%%%%%%%%                      %%%%%%%%%%%%%%%
%%%%%%%%%%%%%%%%%%%%%%%%%%%%%%%%%%%%%%%%%%%%%%%%%%%%
%%%%%%%%%%%%%%%%%%%%%%%%%%%%%%%%%%%%%%%%%%%%%%%%%%%%
%%%%%%%%%%%%%%%%%%%%%%%%%%%%%%%%%%%%%%%%%%%%%%%%%%%%
%%%%%%%%%%%%%%%%%%%%%%%%%%%%%%%%%%%%%%%%%%%%%%%%%%%%
%%%%%%%%%%%%%%%%%%%%%%%%%%%%%%%%%%%%%%%%%%%%%%%%%%%%

\section{Experimental Results}\label{Sec:Simulations}
\begin{figure*}[t]
    \includegraphics[width=\linewidth]{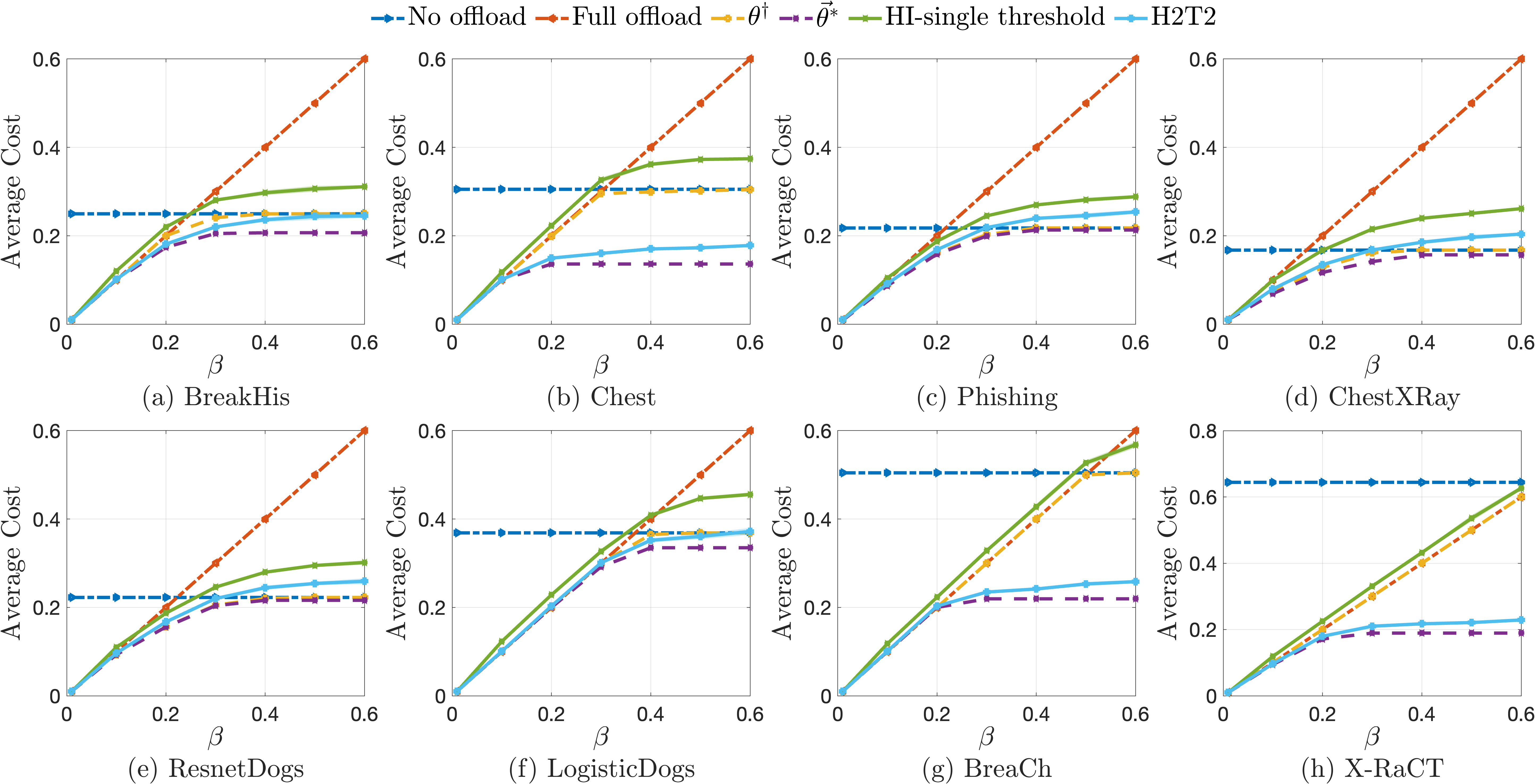}
    \caption{Average cost of H2T2 policy vs. fixed offloading cost $\beta$ for different datasets. Figures (a)-(d) are for in-distribution data, and (e) is for OOD data.}
    \label{fig:perfComparison}
\end{figure*}
\begin{figure*}[t]
    \includegraphics[width=\linewidth]{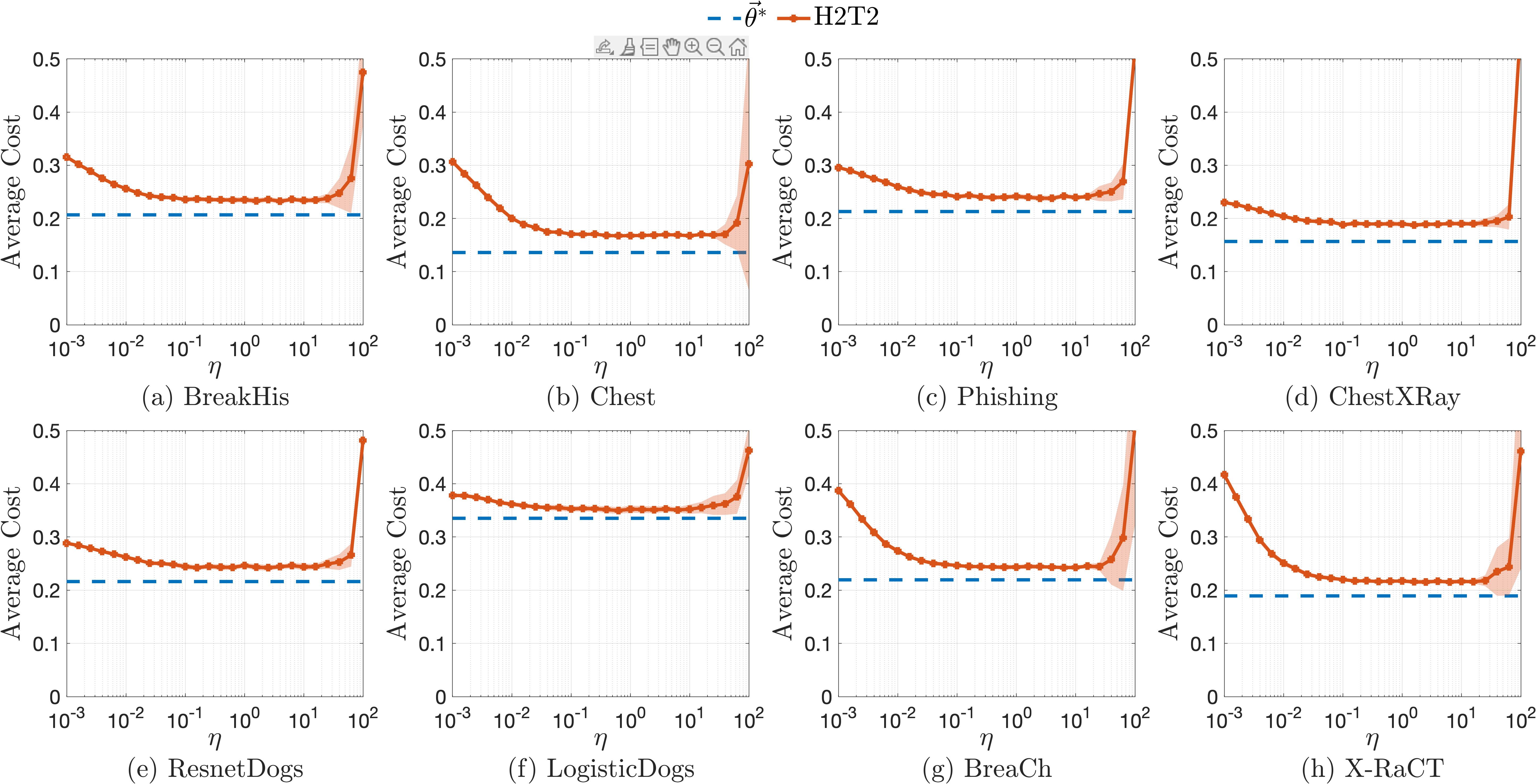}
    \caption{Average cost vs. learning rate $\eta$ with $\beta\!=\!0.4, \delta_1\!=\!0.7,\delta_{-1}\!=\!1, T\!=\!10000$. Similar trends were observed throughout the range of $\beta\in (0,1)$.}
    \label{fig:costvseta}
\end{figure*}
\begin{figure*}[h]
    \includegraphics[width=\linewidth]{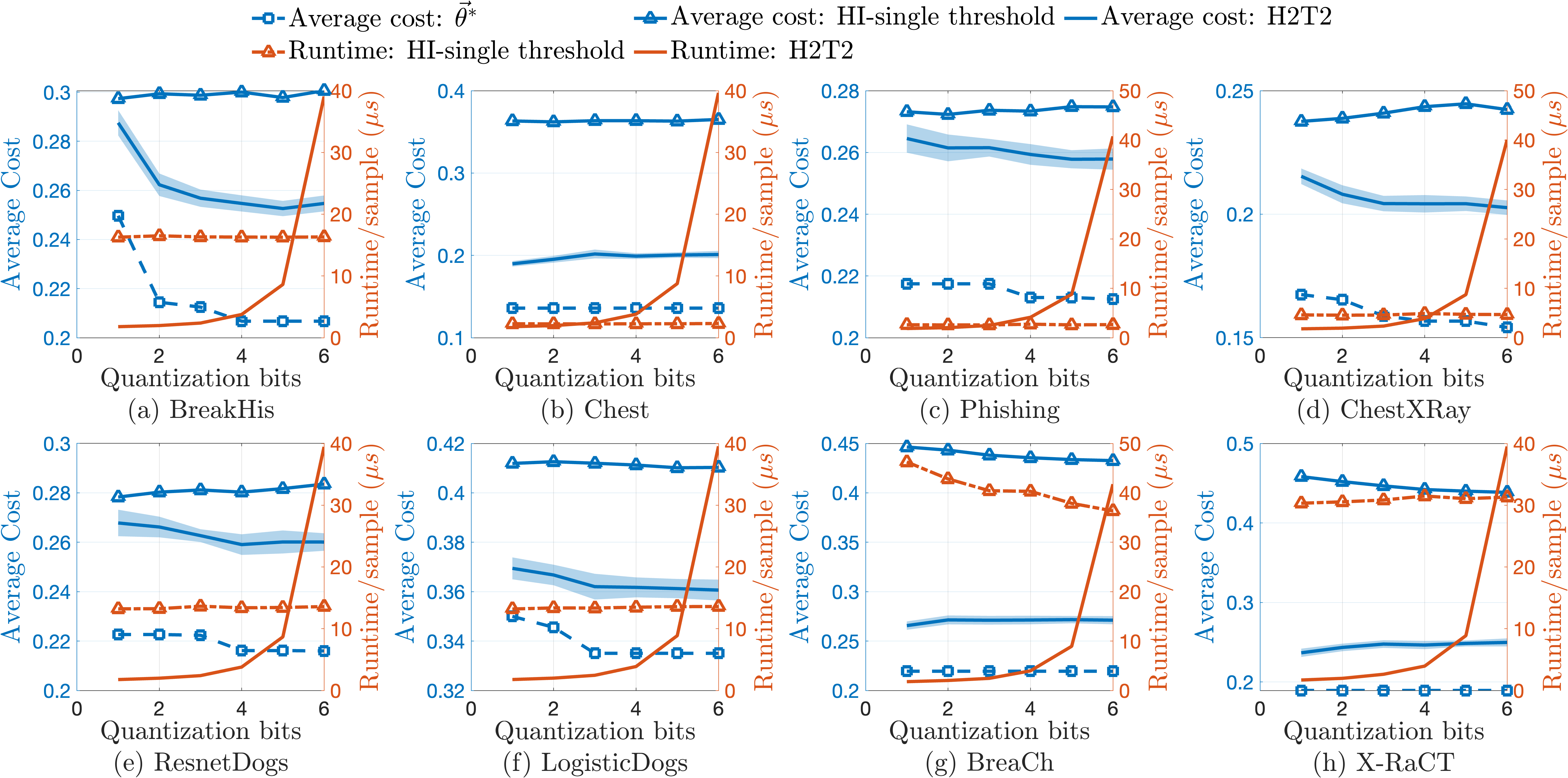}
    \caption{Average cost and runtime-complexity with LDL output quantization}
    \label{fig:runtimecomplexity}
\end{figure*}

We evaluate the performance of H2T2 via experiments on the following dataset-model pairs, which we simply refer to by the dataset name.
\begin{enumerate}
\item The Breast Cancer Histopathological Image Classification \textbf{(BreakHis)}~\cite{breakhis} contains 7909 images of benign (class 1) and malignant cells. The LDL model is a lightweight MobileNet-based classifier with additional dense and dropout layers.
\item \textbf{Chest}~\cite{chestCT} contains 844 chest CT scan images re-categorised into healthy and cancerous (class 1) classes for our use. 
The LDL is based on MobileNet, with a size of 13.3 MB including trainable parameters.
\item\textbf{Phishing}~\cite{phishingDataset,tan2018phishing} includes data collected from 5000 phishing and as many legitimate websites, with additional samples for testing. A logistic regression model (56 bytes) is used as LDL.
\item For the \textbf{Synthetic} dataset, we generate softmax-like values using Gaussian mixtures truncated to $(0,1)$.
\item \textbf{ChestXRay}\cite{chestxray} consists of 5,216 training X-ray images (1,341 normal and 3,875 pneumonia, class1), 624 test images (234 normal and 390 pneumonia), and 16 validation samples. The LDL is a lightweight convolutional neural network with three convolutional layers, max pooling, and global average pooling trained for binary classification.
\item \textbf{ResnetDogs} uses a subset of CIFAR-10~\cite{cifar10,CifarDS} containing only cat and dog images, with 10,000 samples (80\% train, 20\% validation) and 2,000 test images equally divided between the two classes. The LDL is a compact ResNet-8 model with four residual blocks, batch normalization, and global average pooling trained for binary classification.
\item \textbf{LogisticDogs} uses the same CIFAR-10 subset as above but employs a logistic regression model as the LDL. The model, occupying 97~KB of memory, is trained on flattened and standardized image pixels using a Scikit-learn pipeline.
\item\textbf{BreaCh} is created by classifying all 7909 BreakHis samples using the model trained on Chest CT scan. It mimics an {OOD} data scenario with a domain shift for the model trained on the Chest CT scan. As a result, unbeknownst to the user, the accuracy drops below 50\%, resulting in the misdiagnosis of 38\% cancerous tumours.
\item \textbf{X-RaCT} is created by classifying all 5,856 X-ray images from~\cite{chestxray} using the model trained on the chest CT dataset~\cite{chestCT}. Similar to BreaCh, this dataset mimics an {OOD} scenario, where the accuracy drops significantly below that of a random predictor. Although inverting the output could yield better-than-random results, we assume this mismatch is unknown to the user. The dataset is primarily used to illustrate model, data, or distribution shift.
\end{enumerate}
These are summarised in Table~\ref{Table:Dataset}. 
These datasets and models are chosen to span diverse domains that have asymmetric misclassification costs and also show an OOD application. For benchmarking, we implement the following policies:
\begin{enumerate}
\item\textit{No Offload}, where the LDL inference is accepted as is;
\item\textit{Full Offload}, where all samples are offloaded;
\item\textit{HI-single threshold}, the state-of-the-art continuum expert HI policy~\cite{hiTMLCN};
\item$\theta^\dagger$, offline optimal single-threshold policy;
\item$\vec{\theta}^*$, offline optimal two-threshold policy; 
\item\textit{H2T2}, our proposed policy.
\end{enumerate}
We set $\delta_1\!=\!0.7$ and $\delta_{-1}\!=\!1$ and use a fixed offloading cost $\beta$ for comparison study. Similar trends are observed for varying $\delta_1$ and $\delta_{-1}$.
The quantised threshold resolution is set to $0.0625$, resulting in $\vert\Theta\vert\!<\!256$. An increase in resolution did not result in a considerable reduction in average cost for the datasets studied. 
Since the bound optimising parameters from \ref{corol2} do not necessarily minimise regret, we use $\eta^*$ from Corollary~\ref{corol2} and set learning rate $\eta\!=\!1$ without further parameter tuning. 
For fair comparison, $T\!=\!10^4$ samples from each dataset via uniform sampling.
Standard deviation for HI-single threshold and H2T2 are included in all figures, but are negligible (25 algorithm runs). 
%%%%%%%%%%%%%%%%%%%%%%%%%% Dataset as tables%%%%%%%%%%%%%%%%%%%%%%%%%

\subsection{Performance Comparison}\label{subsec:perfAnalysis}
In Fig.~\ref{fig:perfComparison}, we present the average cost incurred by different policies by varying the fixed offloading cost $\beta\!\in\!(0,0.6)$.
We observe that H2T2 outperforms the state-of-the-art HI single-threshold policy and provides a percentage reduction of up to 55\% (for Fig.~\ref{fig:perfComparison}(d) at $\beta\!=\!0.6$) for the dataset, model, parameters and range of $\beta$ used in this illustration.
However, for Phishing in Fig. \ref{fig:perfComparison}(c), the improvement is marginal. This is attributed to the policies yielding near-trivial results, as indicated by the overlap of their respective offline optima and the No-offload policy.
Interestingly, in two datasets, H2T2 outperforms $\theta^\dagger$, the offline optimal single-threshold policy.

For some $\beta$ values, the static polocies outperform online learning policies (e.g., No offload policy with $\beta\!>\!0.3$ in Phishing). However, these regions are dataset-and model-dependent and unknown a priori.

Note the significant cost reduction achieved by H2T2 for BreaCh in  Fig.~\ref{fig:perfComparison}(e) for the OOD data. 
Being model-agnostic and robust to distribution shifts, H2T2 limits the additional cost of data (or model) mismatch to about $0.08$ (or $0.01$) at $\beta\!=\!0.6$; compare Fig.~\ref{fig:perfComparison}(e) with (b) (or (b)). 

We observed that H2T2 learns and adapts to reduce the FNs even when the offload is relatively costly ($\beta\!=\!0.6$). For BreakHis, we observed FNs reducing to 9.5\% with FPs increasing to 14\%.In the OOD setting, we observed H2T2 reducing FNs to below 1\% with FPs increasing to 28\%. 
See the FNs and FPs of the LDL in Table \ref{Table:Dataset}.

In Fig. \ref{fig:costvseta} we plot average cost vs. learning rate $\eta$ with $\beta\!=\!0.4, \delta_1\!=\!0.7,\delta_{-1}\!=\!1, T\!=\!10000$. 
As we have described already, $\eta^*$ from Corollary~\ref{corol2} that minimizes the worst-case regret bound, is not necessarily the one that provides minimum regret in a given setting. We use this figure to illustrate this. Further, while we had chosen $\eta\!=\!1$ t for simplicity without any parameter search, we could see with Fig.~\ref{fig:costvseta} that this choice is rather a good one.
Similar trends were observed throughout the range of $\beta\in (0,1)$.

Finally, in Fig.~\ref{fig:runtimecomplexity}, we plot average cost and runtime-complexity with LDL output quantization. We show the variation of average cost (left y-axis) and runtime (right y-axis) with the quantization in bits for different datasets. A $b$-bit quantization corresponds to a $2^{-b}$ resolution and number of experts $|\Theta|=2^{b-1}(2^b+1)$ after ignoring the invalid region. As a result of the quantization, note that for smaller $b$, H2T2 performs faster than the single-threshold HI that employs a continuum expert without any quantization. However, as $b$ increases, H2T2 takes more time to run in comparison to the state-of-the-art single threshold HI; an expected behavior due to the increase in the dimensionality of the expert space. However, also note that H2T2 often does not require too small a resolution for the datasets studied, as observed from the flat average cost regions. We see a good trade-off at $b=4$, where the complexity is smaller than or comparable to the HI-single threshold, and the average cost has attained a fair amount of stability. This is the reason we used $b=4$ throughout this work.
    
%%%%%%%%%%%%%%%%%%%%%%%%%%%%%%%%%%%%%%%%%%%%%%%%%%%%
%%%%%%%%%%%%%%%%%%%%%%%%%%%%%%%%%%%%%%%%%%%%%%%%%%%%
%%%%%%%%%%%%%%%%%%%%%%%%%%%%%%%%%%%%%%%%%%%%%%%%%%%%
%%%%%%%%%%%%%%%%%%%%%%%%%%%%%%%%%%%%%%%%%%%%%%%%%%%%
%%%%%%%%%%%%%%%%%%%%%%%%%%%%%%%%%%%%%%%%%%%%%%%%%%%%
%%%%%%%%%%%%%%%                      %%%%%%%%%%%%%%%
%%%%%%%%%%%%%%%        SECTION       %%%%%%%%%%%%%%%
%%%%%%%%%%%%%%%                      %%%%%%%%%%%%%%%
%%%%%%%%%%%%%%%%%%%%%%%%%%%%%%%%%%%%%%%%%%%%%%%%%%%%
%%%%%%%%%%%%%%%%%%%%%%%%%%%%%%%%%%%%%%%%%%%%%%%%%%%%
%%%%%%%%%%%%%%%%%%%%%%%%%%%%%%%%%%%%%%%%%%%%%%%%%%%%
%%%%%%%%%%%%%%%%%%%%%%%%%%%%%%%%%%%%%%%%%%%%%%%%%%%%
%%%%%%%%%%%%%%%%%%%%%%%%%%%%%%%%%%%%%%%%%%%%%%%%%%%%

\section{Discussion: Multiclass Classification}
We have thus far focused on cost-sensitive binary classification. We now briefly examine extending our setting to multiclass classification with asymmetric pairwise misclassification costs. The overarching objective remains the same: to partition the softmax value space into distinct regions for each class and an additional region for offload.

Let $K$ be the number of classes and $C \in [0,1]^{K \times K}$ be the normalized cost matrix where $C_{ij}$ is the cost of misclassifying class $i$ as class $j$, and $C_{ii} = 0,\,\forall i$.
Let $f$ be the $K\times1$ softmax vector and $C_k$ denote the $k^{\text{th}}$ column of $C$, representing the cost vector when class $k$ is the prediction.
The loss function $l_t$ follows \eqref{eq:lossFn}, with a modified $\phi_t$ defined as:
\begin{align}
    \phi_t = C_{ij},\quad \text{ if class $i$ is predicted as $j$.}
\label{eq:Phi_t_multiclass}
\end{align}
\begin{theorem}\label{thm:optimal_calibrated_multiClass}
The optimal predictor $h_l^*(x_t)$ of a calibrated LDL designed for a $K$-class classification task is given by: 
\begin{align}
 h_l^*(x_t) = \arg\min_k f^TC_k. \label{eq:Thm1_OptPredictor_multiclass}
\end{align}
Further, the optimum offloading decision for this setup is to offload the sample when $\min_k f^TC_k>\beta_t$
with an associated expected cost of $\mathbb{E}[l_t]=\min\{\beta_t,\min_k f^TC_k\}$.
\end{theorem}

\begin{figure}[t]
\centering
\includegraphics[width=1\linewidth]{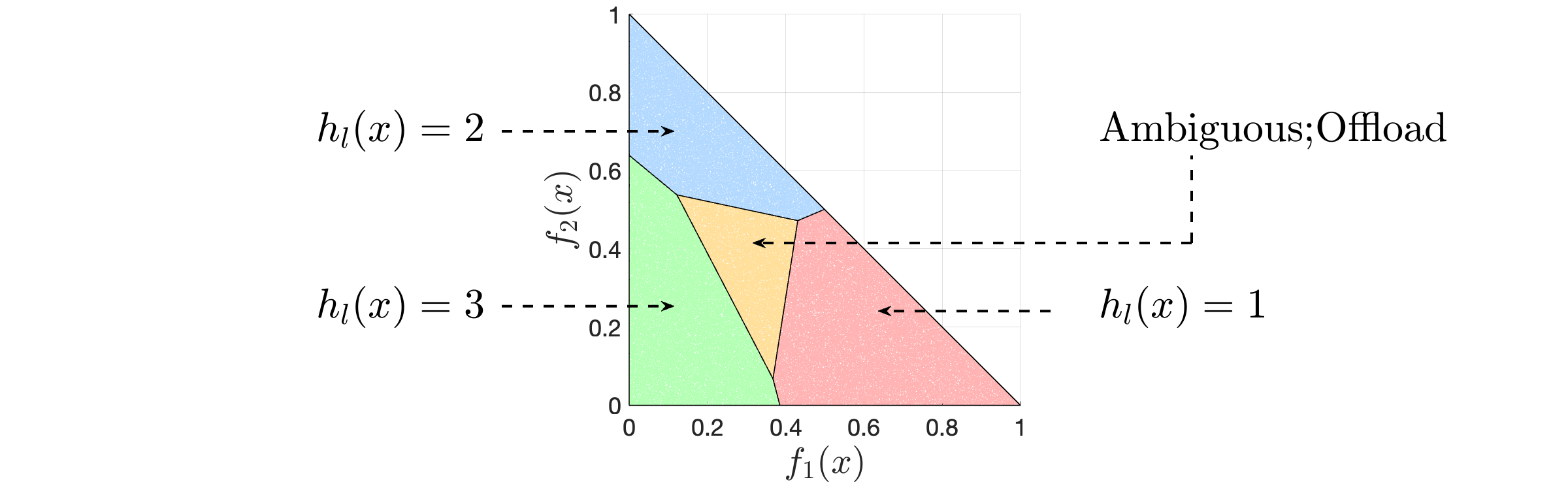}
\caption{Illustration showing the four regions for a three-class classification by a calibrated LDL, plotted in a plane defined by the (first) two independent softmax components. Linear boundaries bound the regions, and the inferences are marked within. $\delta_{1}\!=\!0.7,\delta_{-1}\!=\!1,\beta\!=\!0.4$. A suitable cost matrix $C$ is selected from a random sample for clarity.} 
\label{fig:illustration3Class}
\end{figure}

For calibrated models, this yields $K+1$ decision regions bounded by $(K\!-\!2)$-dimensional simplexes in closed-form. 
For instance, in the binary case ($K\!=\!2$) of previous sections, we saw three regions bounded by 2 points; for $K \!=\! 3$, there are four regions bounded by line segments as illustrated in Fig.~\ref{fig:illustration3Class}. 
Note that we could represent these regions in $K$-dimensions on the $K\!-\!1$-dimensional standard (probability) simplex $f^T\vec{1}=1$, or remove an independent variable and represent it directly in $K\!-\!1$ dimensions.

In the case of uncalibrated models, the boundaries of these regions are not known a priori. 
Similar to the binary setting, we consider a class of experts for each of these boundaries, and design a similar PEA policy to learn them. 
Designing a compact and scalable methodology that utilises the special structure of costs to determine the costs of all experts in a computationally efficient manner is an open problem.

%%%%%%%%%%%%%%%%%%%%%%%%%%%%%%%%%%%%%%%%%%%%%%%%%%%%
%%%%%%%%%%%%%%%%%%%%%%%%%%%%%%%%%%%%%%%%%%%%%%%%%%%%
%%%%%%%%%%%%%%%%%%%%%%%%%%%%%%%%%%%%%%%%%%%%%%%%%%%%
%%%%%%%%%%%%%%%%%%%%%%%%%%%%%%%%%%%%%%%%%%%%%%%%%%%%
%%%%%%%%%%%%%%%%%%%%%%%%%%%%%%%%%%%%%%%%%%%%%%%%%%%%
%%%%%%%%%%%%%%%                      %%%%%%%%%%%%%%%
%%%%%%%%%%%%%%%        SECTION       %%%%%%%%%%%%%%%
%%%%%%%%%%%%%%%                      %%%%%%%%%%%%%%%
%%%%%%%%%%%%%%%%%%%%%%%%%%%%%%%%%%%%%%%%%%%%%%%%%%%%
%%%%%%%%%%%%%%%%%%%%%%%%%%%%%%%%%%%%%%%%%%%%%%%%%%%%
%%%%%%%%%%%%%%%%%%%%%%%%%%%%%%%%%%%%%%%%%%%%%%%%%%%%
%%%%%%%%%%%%%%%%%%%%%%%%%%%%%%%%%%%%%%%%%%%%%%%%%%%%
%%%%%%%%%%%%%%%%%%%%%%%%%%%%%%%%%%%%%%%%%%%%%%%%%%%%

\section{Conclusion}\label{Sec:conclusion}
We consider the task of cost-sensitive binary classification using a local model that works in conjunction with a more accurate but costly remote model. We use the HI framework, where each sample is first sent to the local model, and the decision to offload to the remote model is made based on the output of the local model. 
%Under HI, we propose using a two-threshold rule for local inference and offloading decisions. 
For calibrated local classifiers, we derived a Bayes optimal two-threshold rule. For general, uncalibrated models, we developed H2T2, an online learning policy that finds the optimal thresholds with partial feedback and has sublinear regret. 
Unlike previous HI learning algorithms that use a single threshold for uniform misclassification costs, H2T2 uses two thresholds to handle asymmetric costs and creates an ambiguity region for the offloading decision. It also makes its own cost-sensitive local inferences. H2T2 is a novel approach because it operates during the inference phase without requiring LDL retraining to improve the classification accuracy. It is also agnostic to the models being used. Simulations across various datasets and models show that H2T2 consistently outperforms naive, single-threshold, and even offline single-threshold policies. 
While certain dataset-model combinations might favour naive policies, these conditions are unknown beforehand and are vulnerable to OOD data. H2T2 is a robust and flexible solution for OOD data, making it ideal for real-world applications.

%%%%%%%%%%%%%%%%%%%%%%%%%%%%%%%%%%%%%%%%%%%%%%%%%%%%
%%%%%%%%%%%%%%%%%%%%%%%%%%%%%%%%%%%%%%%%%%%%%%%%%%%%
%%%%%%%%%%%%%%%%%%%%%%%%%%%%%%%%%%%%%%%%%%%%%%%%%%%%
%%%%%%%%%%%%%%%%%%%%%%%%%%%%%%%%%%%%%%%%%%%%%%%%%%%%
%%%%%%%%%%%%%%%%%%%%%%%%%%%%%%%%%%%%%%%%%%%%%%%%%%%%
%%%%%%%%%%%%%%%                      %%%%%%%%%%%%%%%
%%%%%%%%%%%%%%%        SECTION       %%%%%%%%%%%%%%%
%%%%%%%%%%%%%%%                      %%%%%%%%%%%%%%%
%%%%%%%%%%%%%%%%%%%%%%%%%%%%%%%%%%%%%%%%%%%%%%%%%%%%
%%%%%%%%%%%%%%%%%%%%%%%%%%%%%%%%%%%%%%%%%%%%%%%%%%%%
%%%%%%%%%%%%%%%%%%%%%%%%%%%%%%%%%%%%%%%%%%%%%%%%%%%%
%%%%%%%%%%%%%%%%%%%%%%%%%%%%%%%%%%%%%%%%%%%%%%%%%%%%
%%%%%%%%%%%%%%%%%%%%%%%%%%%%%%%%%%%%%%%%%%%%%%%%%%%%

\appendix
\section{Proofs}
\subsubsection{Proof of Theorem~\ref{thm:optimal_calibrated} on page~\pageref{thm:optimal_calibrated}:}
Define an optimal predictor as the one that minimises the expected LDL cost. Let $Y$ denote random variable $h_r(X)$. Recall $\phi_t$ from \eqref{eq:Phi_t} and $f_t\!\coloneqq\!f_1(x_t)$.
{\small\begin{align*}
%     \phi_t &=
% \begin{cases}
% \delta_1 & \text{if FP, i.e., } h_l(x_t) = 1, y_t = 0, \\
% \delta_{-1} & \text{if FN, i.e., } h_l(x_t) = 0, y_t = 1, \\
% 0 & \text{otherwise.}
% \end{cases}\\
    \mathbb{E}[\phi_t]
    % &=\mathbb{E}_X\left[\mathbb{E}_{Y|X}\left[\phi_t\right]\right]\\ 
    &=\mathbb{E}_X\big[\delta_1\indicator_{h_l(X) = 1}\mathbb{P}(Y = 0 \mid X)\\&\qquad+\delta_{-1}\indicator_{h_l(X) = 0}\mathbb{P}(Y = 1 \mid X)\big],\\
    % &=\mathbb{E}_X\big[\delta_1\indicator_{h_l(X) = 1}f_0(X)+ \delta_{-1}\indicator_{h_l(X) = 0}f_1(X)\big],\\
    &=\mathbb{E}_X\big[\delta_1\indicator_{h_l(X) = 1}(1-f(X))+ \delta_{-1}\indicator_{h_l(X) = 0}f(X)\big],
\end{align*}} 
where $\indicator_.$ is the indicator function.
It follows that the optimal predictor of the calibrated LDL, denoted by $h_l^{*}(X)$, activates the indicator function that minimises $\mathbb{E}[\phi_t]$. That is,
{\small\begin{align*}
 h_l^*(X) =
\begin{cases} 
1, & \text{if } 
% \delta_{1}(1-f(X))\leq \delta_{-1}f(X)\\&
\quad\Rightarrow f(X) \geq \tfrac{\delta_1}{\delta_1 + \delta_{-1}}, \\
0, & \text{otherwise},
\end{cases}
\end{align*}}
as given in \eqref{eq:Thm1_OptPredictor}. Thus, with this optimum predictor, we get:  $$\mathbb{E}[\phi_t]=\min\{\delta_{1}(1-f_t),\delta_{-1}f_t\}.$$

Now, let $\mathbb{I}_t$ denote the indicator random variable denoting the decision to offload sample $x_t$.
Then we have the cost $l_t$:  
{\small\begin{equation*}
l_t=\beta_t\mathbb{I}_t+\phi_t(1-\mathbb{I}_t)
\end{equation*}} 
with an the optimum decision is to offload if $\beta_t<\mathbb{E}[\phi_t]$. 
That is, for calibrated LDL with an optimal predictor, the optimum decision is to offload when
{\small\begin{eqnarray*}
\beta_t<\min\{\delta_{1}(1-f_t),\delta_{-1}f_t\},\\
\Rightarrow {\theta_l^*}^{(t)}=\tfrac{\beta_t}{\delta_{-1}}<f_t<1-\tfrac{\beta_t}{\delta_{1}}={\theta_u^*}^{(t)},
\end{eqnarray*}}
with an associated expected cost given by \eqref{eq:Thm1_expectedLoss}.
\qed

\subsubsection{Proof of Lemma~\ref{lemma:unbiased} on page~\pageref{lemma:unbiased}:}
Taking expectation of $\hat{l}$ over realizations of $\zeta$ gives 
{\small\begin{align*}
    \mathbb{E}_{\zeta}[\tilde{l}_t(\Vec{\theta})] &= 
\indicator_{f_t \notin [\theta_l, \theta_u]}\cdot\mathbb{P}(\zeta=1) \tfrac{\phi_t}{\epsilon}+ \indicator_{\theta_l< f_t < \theta_u}\cdot\beta_t,\\
&=\indicator_{f_t \notin [\theta_l, \theta_u]}\cdot\phi_t+\indicator_{\theta_l< f_t < \theta_u}\cdot\beta_t={\hat{l}_t(\Vec{\theta})}.
\nonumber
\end{align*}}
\noindent That is, the estimated loss is an unbiased estimator of the the losses incurred by a threshold-based policy.\hfill$\qed$
% \end{proof}

\subsubsection{Proof of Lemma~\ref{lemma:expDifference} on page~\pageref{lemma:expDifference}:} 
Consider the policy choosing $\Vec{\theta}^{(t)}$ in round $t$, with losses $l_t(\Vec{\theta}^{(t)})$. As before, we overload the notation for easiness and use $l_t(\Vec{\theta})$. We have,
{\small\begin{align}
    \mathbb{E}\left[ L_T \right] &= \sum_{t=1}^{T} \sum_{\Vec{\theta} \in \Theta} \tfrac{w_t(\Vec{\theta})}{W_t} \mathbb{E}_{\zeta_t} \left[ l_t(\Vec{\theta}) \right],\label{elt}\\
    \mathbb{E} \left[ {\hat{L}_T} \right] &= \sum_{t=1}^{T} \sum_{\Vec{\theta} \in \Theta} \tfrac{w_t(\Vec{\theta})}{W_t} \mathbb{E}_{\zeta_t} \left[ {\hat{l}_t(\Vec{\theta})} \right].
    \label{elbart}
\end{align}}
Now, from \eqref{lossdefinition} and \eqref{eq:h2t2trueloss},
{\small\begin{align}
{\hat{l}_t(\Vec{\theta})}&=\indicator_{f_t \notin [\theta_l, \theta_u]}\phi_t+\left(1-\indicator_{f_t \notin [\theta_l, \theta_u]}\right)\beta_t,\nonumber\\
% \intertext{and}
    l_t(\Vec{\theta})&= \indicator_{f_t \notin [\theta_l, \theta_u]} \indicator_{\zeta_t = 0}\phi_t
    + \left( 1 - \indicator_{f_t \notin [\theta_l, \theta_u]} \indicator_{\zeta_t = 0}\right)\beta_t. \nonumber
\end{align}}
\noindent Taking expectation and combining, we get,
{\small\begin{align}
% \mathbb{E}_{\zeta}\!\left[l_t(\Vec{\theta})\right]\!=\!
% \indicator_{f_t \notin [\theta_l, \theta_u]}(1\!-\!\epsilon)\phi_t
%     \!+\! \left(1\!-\!\indicator_{f_t \notin [\theta_l, \theta_u]}(1\!-\!\epsilon)\right)\!\beta.\\
% \mathbb{E}_{\zeta} \left[ l_t(\Vec{\theta}) \right] - l_t(\Vec{\theta}) &= \epsilon \indicator_{f_t \notin [\theta_l, \theta_u]}(\beta_t -\phi_t).\nonumber\\
% \Rightarrow 
\mathbb{E}_{\zeta} \left[ l_t(\Vec{\theta}) \right] - \mathbb{E}_{\zeta} \left[ {\hat{l}_t(\Vec{\theta})} \right] &= \epsilon \indicator_{f_t \notin [\theta_l, \theta_u]}(\beta_t -\phi_t)\leq\epsilon\beta_t.\nonumber
\end{align}}
In the last step, we used Lemma \ref{lemma:unbiased}. Since this holds for any $\Vec{\theta}$ and $t$, we have for any policy and time horizon $T$
{\small\begin{align*}
 &\mathbb{E} \left[ L_T \right] - \mathbb{E} \left[ {\hat{L}_T}\right]\leq\epsilon\beta_t T \leq\epsilon\beta T\leq \epsilon T.\tag*{\qed}
\end{align*}}

\subsubsection{Proof of Theorem~\ref{thm:regret} on page~\pageref{thm:regret}:}
Recall $\tilde{L}_T(\Vec{\theta})$: the cumulative estimated loss with a fixed $\Vec{\theta}$. Consider  $\ln \tfrac{W_{T+1}}{W_1}$:
{\small\begin{align}
\ln \tfrac{W_{T+1}}{W_1} &=  \ln \sum_{\Vec{\theta} \in \Theta} e^{-\eta \sum_{t=1}^{T} \tilde{l}_{t}(\Vec{\theta})} - \ln(|\Theta|), \notag
% \\ &\geq \ln \max_{\Vec{\theta} \in \Theta} e^{-\eta \tilde{L}_T(\Vec{\theta})} - \ln(|\Theta|), \notag
% \\ &\geq \max_{\Vec{\theta} \in \Theta} -\eta L_T(\Vec{\theta}) - \ln(|\Theta|) \notag
\\ &\geq -\min_{\Vec{\theta} \in \Theta} \eta \tilde{L}_T(\Vec{\theta}) - \ln(|\Theta|). \label{-min}
\end{align}}
Using Lemma~\ref{lemma:unbiased} and Jensen's inequality on expectations:
{\small\begin{align}
    \mathbb{E}_{\zeta} \left[ \ln \tfrac{W_{T+1}}{W_1} \right] &\geq -\min_{\Vec{\theta} \in \Theta} \eta \mathbb{E}_{\zeta} \left[ \tilde{L}_T(\Vec{\theta}) \right] - \ln(|\Theta|) \notag \\
   &= -\min_{\Vec{\theta} \in \Theta} \eta {\hat{L}_T(\Vec{\theta})} - \ln(|\Theta|). \label{log_w}
\end{align}}
Next, we upper bound $\log \tfrac{W_{T+1}}{W_1}$ as follows:
{\small\begin{align}
\ln &\tfrac{W_{t+1}}{W_t}
% =\ln \sum_{\Vec{\theta} \in \Theta} \tfrac{w_{t+1}(\Vec{\theta})}{W_t}\notag\\ &
= \ln \sum_{\Vec{\theta} \in \Theta} \tfrac{w_t(\Vec{\theta})}{W_t} e^{-\eta \tilde{\ell}_t(\Vec{\theta})} \notag\\ 
&\leq \ln \sum_{\Vec{\theta} \in \Theta} \tfrac{w_t(\Vec{\theta})}{W_t} \left( 1 - \eta \tilde{\ell}_t(\Vec{\theta}) + {\left(\eta \tilde{\ell}_t(\Vec{\theta})\right)^2}/{2} \right) \label{line2}
\\&= \ln \left( 1 + \sum_{\Vec{\theta} \in \Theta} \tfrac{w_t(\Vec{\theta})}{W_t}
\left( 
-\eta \tilde{\ell}_t(\Vec{\theta}) + {\left(\eta \tilde{\ell}_t(\Vec{\theta})\right)^2}/{2} 
\right) 
\right). \notag
\end{align}}
Simplifying, we get,
{\small\begin{align}
\ln \tfrac{W_{t+1}}{W_t}&\leq \sum_{\Vec{\theta} \in \Theta} \tfrac{w_t(\Vec{\theta})}{W_t} \left( -\eta \tilde{\ell}_t(\Vec{\theta}) + {\left( \eta \tilde{\ell}_t(\Vec{\theta})\right)^2}/{2} \right) \label{line4}
\\ &\leq \sum_{\Vec{\theta} \in \Theta}\tfrac{w_t(\Vec{\theta})}{W_t} \left( -\eta \tilde{\ell}_t(\Vec{\theta}) + {\eta^2 \tilde{\ell}_t(\Vec{\theta})}/{2\epsilon} \right). \label{line5}
\end{align}}

\noindent Above, we have used $e^{-x} \leq 1 - x + \tfrac{x^2}{2}$, $\ln(1 + x) \leq x$, and $\tilde{\ell}_t(\Vec{\theta})\leq\tfrac{1}{\epsilon}$ in \eqref{line2}, \eqref{line4}, and \eqref{line5}, respectively.
% and $-x+x^2/2\geq-0.5,\,\forall x.$
Now we sum \eqref{line5} telescopically over the entire horizon $T$ as follows:
{\small\begin{align}
    \ln &\tfrac{W_{T+1}}{W_1} = \ln \prod_{t=1}^{T} \tfrac{W_{t+1}}{W_t} = \sum_{t=1}^{T} \ln \tfrac{W_{t+1}}{W_t}, \label{log_sum}\\
      &\leq \sum_{t=1}^{T} \sum_{\Vec{\theta} \in \Theta} \tfrac{w_t(\Vec{\theta})}{W_t} \left( -\eta \tilde{\ell}_t(\Vec{\theta}) + \tfrac{\eta^2 \tilde{\ell}_t(\Vec{\theta})}{2\epsilon} \right). 
     \label{logw}\\
% \end{align}
% % Now, taking expectation with respect to $\zeta$, we get
% \begin{align}
\Rightarrow\mathbb{E}\Big[\ln &\tfrac{W_{T+1}}{W_1}\Big] \leq -\eta \sum_{t=1}^{T} \sum_{\Vec{\theta} \in \Theta} \tfrac{w_t(\bar{\theta)}}{W_t} \mathbb{E}_{\zeta}[ \tilde{l}_t(\Vec{\theta})]  \notag
\\ &\qquad\qquad\qquad+ \eta^2\sum_{t=1}^{T} \sum_{\Vec{\theta} \in \Theta} \tfrac{w_t (\Vec{\theta})}{W_t} \tfrac{\mathbb{E}_{\zeta}[ \tilde{l}_t(\Vec{\theta})]}{2 \epsilon} \notag
\\&\leq -\eta \mathbb{E} [{\hat{L}_T(\Vec{\theta})}] + \tfrac{\eta^2}{2 \epsilon} T \label{e_tau}
\\&\leq -\eta \mathbb{E} [L_T(\Vec{\theta})] + \eta \epsilon\beta T + \tfrac{\eta^2}{2 \epsilon} T. \label{tterm}
\end{align}}
\noindent In \eqref{e_tau}, we use $\mathbb{E}_{\zeta}[\tilde{l}_t(\Vec{\theta})]\!=\!{\hat{l}_t(\Vec{\theta})} \!\leq\! 1$ and the expectation is taken over expert selection and $\zeta$. In \eqref{tterm}, we use Lemma~\ref{lemma:expDifference}. The result follows by comparing bounds from \eqref{log_w} and \eqref{tterm}:
{\small\begin{align*}
    \eta \mathbb{E} [L_T(\Vec{\theta})]\!-\!\min_{\Vec{\theta} \in \Theta} \eta {\hat{L}_T(\Vec{\theta})} &\leq  \eta \epsilon\beta T + \tfrac{\eta^2 T}{2\epsilon} + \log(|\Theta|).
\end{align*}}
Now, recall the regret definition \eqref{reg_def}. We get:
{\small\begin{align*}
    R_T & \leq \left(\epsilon\beta + \tfrac{\eta}{2\epsilon}\right)T + \tfrac{\log(|\Theta|)}{\eta}.
\end{align*}}
Corollary~\ref{corol2} follows by setting partial derivatives to zero. 
While not optimal, choosing a simpler $\eta\!=\!\left( 2 \ln^2(|\Theta|)/ T^2 \right)^{{1}/{3}}$ and $\epsilon\!=\!\sqrt{\eta/2}$ also yields regret in the same order: $O(T^{{2}/{3}})$.\qed

\subsubsection{Proof of Theorem~\ref{thm:optimal_calibrated_multiClass} on page~\pageref{thm:optimal_calibrated_multiClass}:}
Let random variables $X,Y$ and $\indicator_.$ denote $x_t, h_r(x_t)$ and the indicator function.
{\small\begin{align*}
    \mathbb{E}[\phi_t]
    % &=\mathbb{E}_X\left[\mathbb{E}_{Y|X}\left[\phi_t\right]\right]\\ 
&=\mathbb{E}_X\sum_{j=1}^K\indicator_{h_l(X) = j}\sum_{i=1}^K\mathbb{P}(Y = i \mid X)C_{ij},\\
&=\mathbb{E}_X\sum_{j=1}^K\indicator_{h_l(X) = j}f^TC_{j}.
\end{align*}}  
It follows that $h_l^*(X)$ activates the $j$ that minimizes $f^T C_j$:
{\small\begin{align*}
h_l^*(X) = \arg\min_j f^TC_j,    
\end{align*}}
\noindent with $\mathbb{E}[\phi_t]\!=\!\min_jf^TC_j$. Now, let $\mathbb{I}_t$ denote the decision to offload $x_t$.
We have, $l_t\!=\!\beta_t\mathbb{I}_t\!+\!\phi_t(1-\mathbb{I}_t).$ Thus, it is optimal to offload if $\beta_t\!<\!\mathbb{E}[\phi_t]$. 
That is, for calibrated LDL with optimal predictor, it is optimal offload when $\beta_t\!<\!\min_k\{f^TC_k\}$,
with $\mathbb{E}[l_t]\!=\!\min\{\beta_t,\min_k f^TC_k\}$.
% \end{proof}
\qed
\small
\bibliography{references}
\end{document}